\documentclass[sigconf,natbib]{acmart}

\usepackage{pgfplots}
\usepackage{wrapfig}

\usepackage{caption}
\usepackage{CJKutf8}
\usepackage{booktabs}
\usepackage{array}
\usepackage{graphicx}
\pgfplotsset{compat=1.16}

\AtBeginDocument{%
  \providecommand\BibTeX{{%
    \normalfont B\kern-0.5em{\scshape i\kern-0.25em b}\kern-0.8em\TeX}}}

\begin{document}
\begin{CJK}{UTF8}{gbsn}

\title{Leveraging Large Language Models for Relevance Judgments in Legal Case Retrieval}

\author{Shengjie Ma}
\affiliation{%
  \institution{Gaoling School of Artificial Intelligence, Renmin University of China}
  \city{Beijing}
  \country{China}}
\email{msj@ruc.edu.cn}

\author{Qi Chu}
\affiliation{%
  \institution{Beijing Normal University}
  \city{Beijing}
  \country{China}}
\email{13521927989@163.com}

\author{Jiaxin Mao}
\authornote{Corresponding authors.}
\affiliation{%
  \institution{Gaoling School of Artificial Intelligence, Renmin University of China}
  \city{Beijing}
  \country{China}}
\email{maojiaxin@gmail.com}

\author{Xuhui Jiang}
\affiliation{%
  \institution{DataArc Tech Ltd.}
  \city{Shenzhen}
  \country{China}}
\email{jiangxuhui@idea.edu.cn}

\author{Haozhe Duan}
\affiliation{%
  \institution{Gaoling School of Artificial Intelligence, Renmin University of China}
  \city{Beijing}
  \country{China}}
\email{duanhaozhe@ruc.edu.cn}

\author{Chong Chen}
\affiliation{%
  \institution{Huawei Cloud BU}
  \city{Beijing}
  \country{China}}
\email{chenchong55@huawei.com}




\renewcommand{\shortauthors}{}

\begin{abstract}
Determining which legal cases are relevant to a given query involves navigating lengthy texts and applying nuanced legal reasoning. Traditionally, this task has demanded significant time and domain expertise to identify key Legal Facts and reach sound juridical conclusions. In addition, existing data with legal case similarities often lack interpretability, making it difficult to understand the rationale behind relevance judgments. With the growing capabilities of large language models (LLMs), researchers have begun investigating their potential in this domain. Nonetheless, the method of employing a general large language model for reliable relevance judgments in legal case retrieval remains largely unexplored. To address this gap in research, we propose a novel few-shot approach where LLMs assist in generating expert-aligned interpretable relevance judgments.
The proposed approach decomposes the judgment process into several stages, mimicking the workflow of human annotators and allowing for the flexible incorporation of expert reasoning to improve the accuracy of relevance judgments. Importantly, it also ensures interpretable data labeling, providing transparency and clarity in the relevance assessment process. Through a comparison of relevance judgments made by LLMs and human experts, we empirically demonstrate that the proposed approach can yield reliable and valid relevance assessments. Furthermore, we demonstrate that with minimal expert supervision, our approach enables a large language model to acquire case analysis expertise and subsequently transfers this ability to a smaller model via annotation-based knowledge distillation.

\end{abstract}




\maketitle

\section{Introduction}
Legal case retrieval (LCR) is a critical component of modern legal systems, for ensuring fairness and justice. It enables legal professionals to find relevant cases related to their current cases. According to "Guiding Opinions of the Supreme People's Court of China on Unified Legal Application and Strengthening Case Retrieval": the term "similar cases" refers to cases that share similarities in basic facts, points of dispute, issues of legal application, and other aspects with the pending case; when the pending case lacks clear judicial rules or has not yet formed unified judicial rules, case retrieval for similar cases should be conducted. 

 Many practitioners still rely on keyword-based retrieval systems when searching for similar cases. However, this approach is cumbersome, has a steep learning curve, and offers low efficiency and accuracy. The researchers have made extensive attempts at the LCR task. BERT-PLI\cite{shao2020bert} models the relevance by aggregating paragraph-level interactions of case pairs. SAILER\cite{li2023sailer} aims to maximize the utilization of information in annotated data. Specifically, it utilizes an encoder-decoder architecture for pre-training, which encodes the facts into vectors and uses a decoder to re-construct the vectors to the original decisions and reasoning. In addition, \citet{wei2022chain} proposes IOT-Match, which learns to find rationales from paired legal cases based on semantics and legal elements of sentences, with the assistance of numerous expert-annotated interpretable relevance labels. Despite the promising results of these models on specific test sets, they still rely heavily on high-quality annotated data. Owing to the scarcity of annotated data, a notable concern arises regarding the limited generalizability of these models across diverse data.
 
\begin{figure}
    \centering
    \includegraphics[width=1\linewidth]{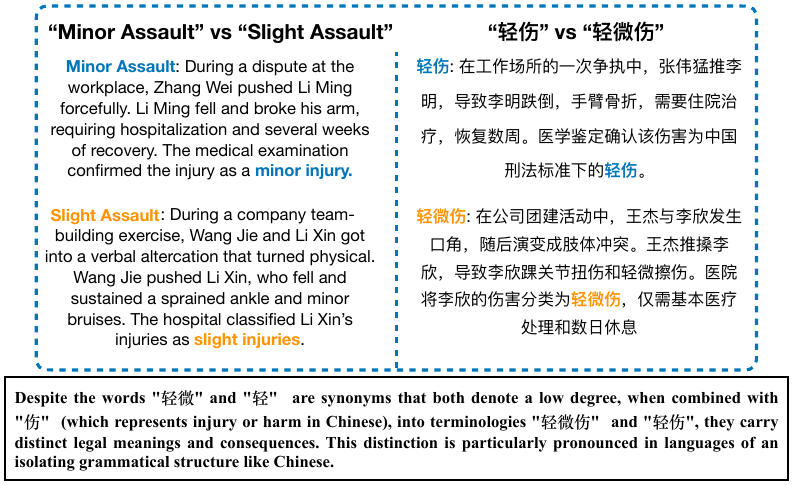}
    \caption{An example of a challenge of legal relevance judgments.}
    \label{fig0}
\end{figure}

These concerns highlight the role of a more reliable and valid method to collect interpretable legal relevance judgments, which is vital not only for the evaluation of models but also for the training of more sophisticated models. 
However, to judge relevance, one necessitates substantial domain expertise and a significant effort to comprehend lengthy texts~\cite{shao2023understanding}. This means that manual judgments require much more complex preliminary instructions for annotators, leading to higher learning effort, greater demands on their abilities, and difficulties in expanding the scale of annotated data. 
In particular, the major challenges in legal cases relevance judgments are threefold: \textbf{1. Expertise-intensive}: Accurate judgments in the legal domain demands high-level domain knowledge. This includes understanding the intrinsic connection between factual scenarios and their corresponding judicial interpretations. It also requires recognizing how objective conduct and subjective attitudes interact to shape the final legal conclusion. \textbf{2. Lengthy-text}: The query and candidates often contain thousands of words. \textbf{3. Nuance-sensitive}: As shown in Figure \ref{fig0}, besides a comprehensive understanding of the overall context, sensitivity to nuances in context and terms is equally important. Because such minor distinctions can significantly impact judicial interpretations and decisions, e.g., "homicide" vs "manslaughter", "minor injury" vs "minor assault", and "robbery" vs "Theft". (Please refer to Section ~\ref{Relevance} Relevance part for further details of legal relevance). 

Existing work\cite{thomas2023large, Faggioli_2023} has shown that the judgment capabilities of large language models (LLMs) in general NLP tasks can rival those of crowd workers. Most efforts often involved general relevance or direct relevance signals, like question-answering, and text classification\cite{thalken2023modeling,gilardi2023chatgpt}. Since relevance labels in legal cases are far more complicated as mentioned earlier, our major research question is whether LLM can perform relevance judgments in the legal field. 
Although researchers have developed legal domain-specific large language models (LLMs) such as ChatLaw\cite{cui2023chatlaw} and LawGPT\cite{lawgpt}, these models are primarily fine-tuned for general legal question-answering tasks. Due to limitations in model parameters and the volume of training data, their understanding and reasoning capabilities are inferior to those of general-purpose LLMs like ChatGPT, making it challenging to directly employ them for relevance judgments in long-form legal cases. Furthermore, ChatLaw is trained mainly on English legal data due to the scarcity of high-quality annotated Chinese legal data\cite{cui2023chatlaw}. This underscores the critical need to develop and curate more high-quality Chinese legal datasets to enhance the performance and applicability of legal models in Chinese legal contexts.

In this work, we will focus on developing an automatic relevance judgment method for legal case retrieval task. Instead of utilizing or building specialized LLMs, which carry many serious constraints in cost and capabilities (Please refer to the next section for details), we employ one of the most popular general LLMs, ChatGPT, for domain-specific judgments. To take advantage of the understanding and reasoning capabilities of the general LLM, we instruct it step-by-step\cite{wei2022chain} with minimal expert guidance. As a result, our approach can enable a general LLM to make judgments on complex legal case relevance in an interpretable manner, closely aligned with those of human experts. Specifically, we address the challenge of \textbf{Lengthy-text} by decomposing the judgment into fine-grained sub-processes, including distinct views in fact extractions and relevance judgments. In each sub-process, we elaborate expert's reasoning process as instructions to ease the \textbf{Expertise-Intensive} challenge and mitigate the \textbf{Nuance-Sensitivity} issue. In addition, we employ adaptive retrieval of effective demonstrations to further tackle the challenge of \textbf{Expertise-Intensive}. Given a group of unlabeled legal cases, we also designed a strategy for efficiently collecting possible positive case pairs.
 
Furthermore, we leveraged label-only synthetic data to fine-tune downstream retrieval models, significantly enhancing the performance of legal case retrieval. Notably, by using synthetic data with both labels and interpretable reasoning generated during the judgment process, we trained two 7B-scale LLMs that match or even outperform GPT-3.5 in judging legal case relevance. These two applications yielded substantial performance gains, providing strong empirical support for the effectiveness of our approach.

We summarize the major contributions of the paper as follows:
\begin{itemize}
    \item We design an automatic workflow to leverage a general LLM to make relevance judgments for legal cases. To the best of our knowledge, this is the first attempt at automated legal case relevance judgments.
    \item We evaluate and analyze the quality of automatic relevance judgments by comparing them with human labels. Empirical experiments demonstrate that our approach can achieve high consistency with human experts.
    \item We use the proposed method to generate a synthetic dataset and show that with minimal expert supervision, our approach enables a large language model to acquire case analysis expertise, and subsequently transfer this ability to a smaller model via annotation-based knowledge distillation. 
    


\end{itemize}



\section{Related works}
\subsection{Pre-trained Language Models (PLMs) in Legal case retrieval}
Pre-trained Language Models (PLMs) like BERT\cite{devlin2018bert}, GPT-2\cite{lagler2013gpt2}, and RoBERTa\cite{liu2019roberta} have set new standards in a variety of NLP tasks. The strength of PLMs lies in their ability to understand context, making them effective for tasks ranging from sentiment analysis to machine translation. However, in the legal domain such as legal case retrieval, general PLMs still struggle with the complex nuances and intensive background domain knowledge of legal jargon and may produce biased or asymmetric predictions. Additionally, legal documents typically consist of thousands of tokens, far exceeding the length that mainstream PLMs can handle\cite{xiao2021lawformer}. Researchers are dedicated to designing specialized models\cite{shao2020bert, li2023sailer, chalkidis2020legal, xiao2021lawformer} tailored to legal case matching, underscoring the limitations of generic PLMs in accurately interpreting and classifying extensive legal content. 

Large-scale Language Models (LLMs) are an extension of the PLM concept but on a much grander scale, such as GPT4\cite{openai2023gpt4} and LLaMa\cite{touvron2023llama}. With billions or even trillions of parameters, LLMs are designed to understand and generate human-like text with little to no fine-tuning required for specific tasks, showing significant potential in various fields, including law. 

Although researchers have attempted to train domain-specific LLMs for the legal field\cite{cui2023chatlaw,lawgpt}, there are still several drawbacks in directly training a dedicated large language model for this domain: 1. The process requires a substantial amount of high-quality data and expensive computational resources. Legal texts can be vast and diverse, necessitating extensive training data to ensure the model's proficiency in understanding the nuances of legal language and reasoning. 2. The legal domain encompasses numerous downstream tasks, each with its own specific requirements and nuances. As a result, domain-specific large-scale language models lack generalizability and reliability across various downstream tasks. Adapting these models to each specific task requires fine-tuning with task-specific data, leading to considerable efforts for collecting specialized training data for each task. 3. Compared to commercial general large models, these domain-specific large models suffer from insufficient training scale and data volume, leading to weaker model memorization, comprehension, and reasoning capabilities. Additionally, most of these domain-specific models have mainly undergone fine-tuning for legal question-answering tasks.

In this article, we present a lightweight approach that enables general LLMs for automated legal relevance judgments.


\subsection{LLM Agents for Data Augmentation}
Using LLM agents for data augmentation is a promising avenue, especially for tasks where data collection is challenging. \cite{wang2022semantic} proposed a method that combines semantic data augmentation with distance metric learning for domain generalization. \cite{meyer2022we} explored the feasibility of GPT-3 generated synthetic data for training conversational AI classifiers, revealing that while synthetic data offers advantages, it doesn't surpass real user data in performance. In addition, \cite{saad2023udapdr} develop a method for using large language models (LLMs) to generate large numbers of synthetic queries cheaply, which generates a small number of synthetic queries using an expensive LLM and then uses a much less expensive one to create large numbers of synthetic queries that are used to fine-tune re-rankers. These studies underscore the growing importance and potential of utilizing LLM to generate synthetic data, especially in scenarios with limited data resources. While it offers a novel way to leverage the capabilities of large-scale language models, care must be taken to ensure that the generated data aligns well with real-world scenarios and doesn't introduce biases or errors. As with any data augmentation technique, it's crucial to validate its effectiveness empirically on the target task. Therefore, in this work, we will evaluate our automatic judgment approach from two levels: the consistency with human labeling and the empirical effectiveness of data-augmented models on the LCR task.

 \section{Methodology}
\begin{figure*}[h]  
	\centering
	\includegraphics[width=2.1\columnwidth]{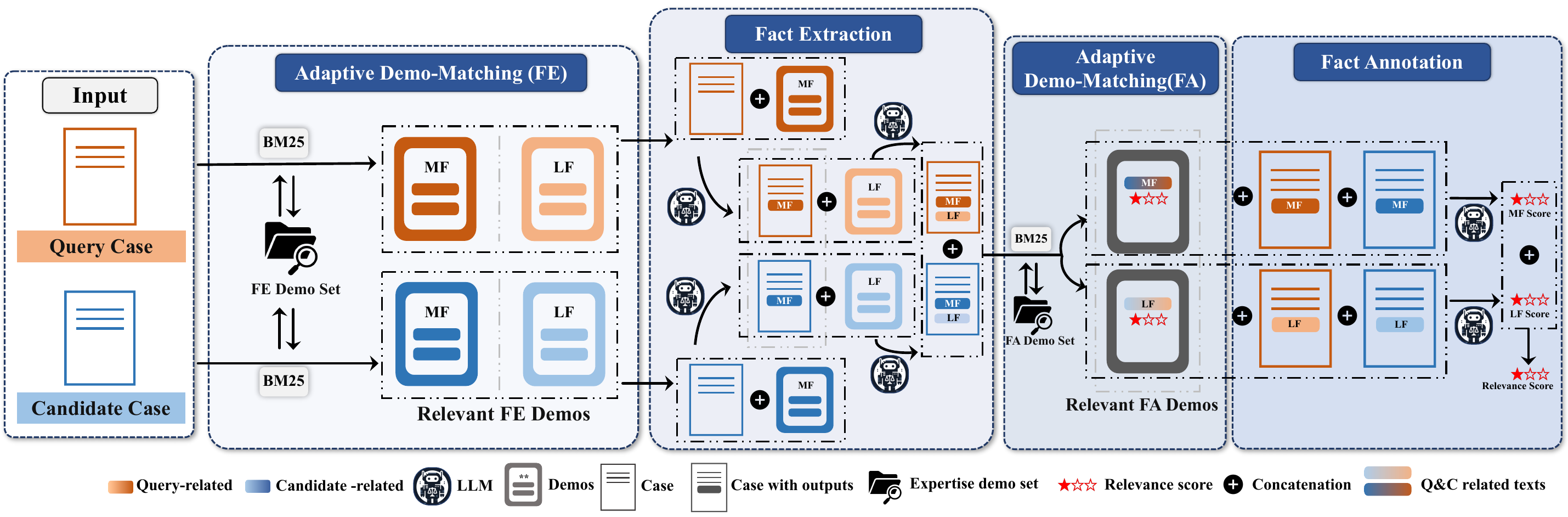} 
	\caption{The proposed workflow of data judgments for legal case retrieval task. To be clear, please note that Adaptive Demo-Matching (ADM) is applied twice before Fact Extraction (FE) and Fact Annotation (FA).}      
	\label{Figure 1}
\end{figure*}
\begin{figure}
    \centering
    \includegraphics[width=1\linewidth]{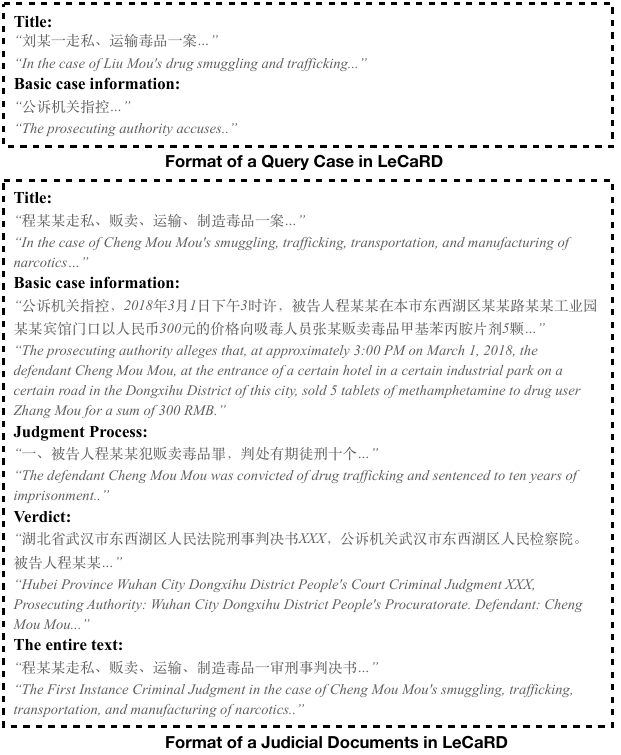}
    \caption{An example of data format in LeCaRD. Note that the original language is Chinese and the English texts are translations.}
    \label{fig_lecard}
\end{figure}

\subsection{Preliminary}

\textbf{Relevance}\label{Relevance}:
According to LeCaRD\cite{ma2021lecard}, the Chinese legal case retrieval dataset that we use, of which the data format is shown in Figure \ref{fig_lecard}, the determination of case relevance primarily depends on the key facts of the case, which are composed of "\textbf{Material Facts (MF)}" and "\textbf{Legal Facts (LF)}". Legal Facts are the legal evaluations of Material Facts. For example, in a hypothetical case:

\textbf{Facts}: A had persistent conflicts with B. Consequently, A found a pretext to provoke B and then assaulted him. B's injuries were classified as 'Level 2 minor injury.

\textbf{Legal Facts}: Arbitrary assault on others; causing minor injuries to others; acts that disrupt social order.

\textbf{Material Facts}: Occasional conflicts, seeking trouble under pretense, unprovoked assault; "Level 2 minor injury."

To determine whether different cases are related, one should deep into the Material Facts and the Legal Facts. The Material Facts of a case include the identity of multiple factors such as the defendant and the victim, the cause, process, outcome, time, and location of the crime committed by the defendant. Judging the relevance of MF involves a "factual level" assessment of the relatedness of different cases, focusing on whether there is "identity" between multiple factors. Judging the relevance of LF includes the subjective motive of the defendant, whether the victim is at fault, and whether there is a causal relationship between the criminal act and the damage caused. This involves an adaptive "legal level" assessment in different legal aspects, focusing on whether there is "homogeneity" between two cases. In practice, the judgment should follow the order of "Material Facts first, Legal Facts second." Both are necessary and sufficient conditions for the assessment of case-relevance.

\textbf{Problem Formulation}: The legal case retrieval task is defined as finding cases related to a given query case from a set of candidate cases. Specifically, given a query case \( q \) and a set of candidate cases \( D \), the objective is to retrieve the top-\( k \) relevant cases from a vast candidate pool, denoted as \( D_q = d_1, d_2, \ldots, d_k \). 

\subsection{Automated Relevance Judgments for Legal Cases}
We have meticulously designed a prompt workflow for legal case retrieval, which can be segmented into four steps: 1) Preliminary factual analysis by legal professionals, 2) \textbf{Adaptive Demo-Matching (ADM)}, 3) \textbf{Fact Extraction (FE)}, and 4) \textbf{Fact Annotation (FA)}. Figure \ref{Figure 1} shows the overall judgment workflow. In the subsequent sections, we will delve into each of these components in detail.

\begin{figure}[h]  
	\centering
	\includegraphics[width=1\columnwidth]{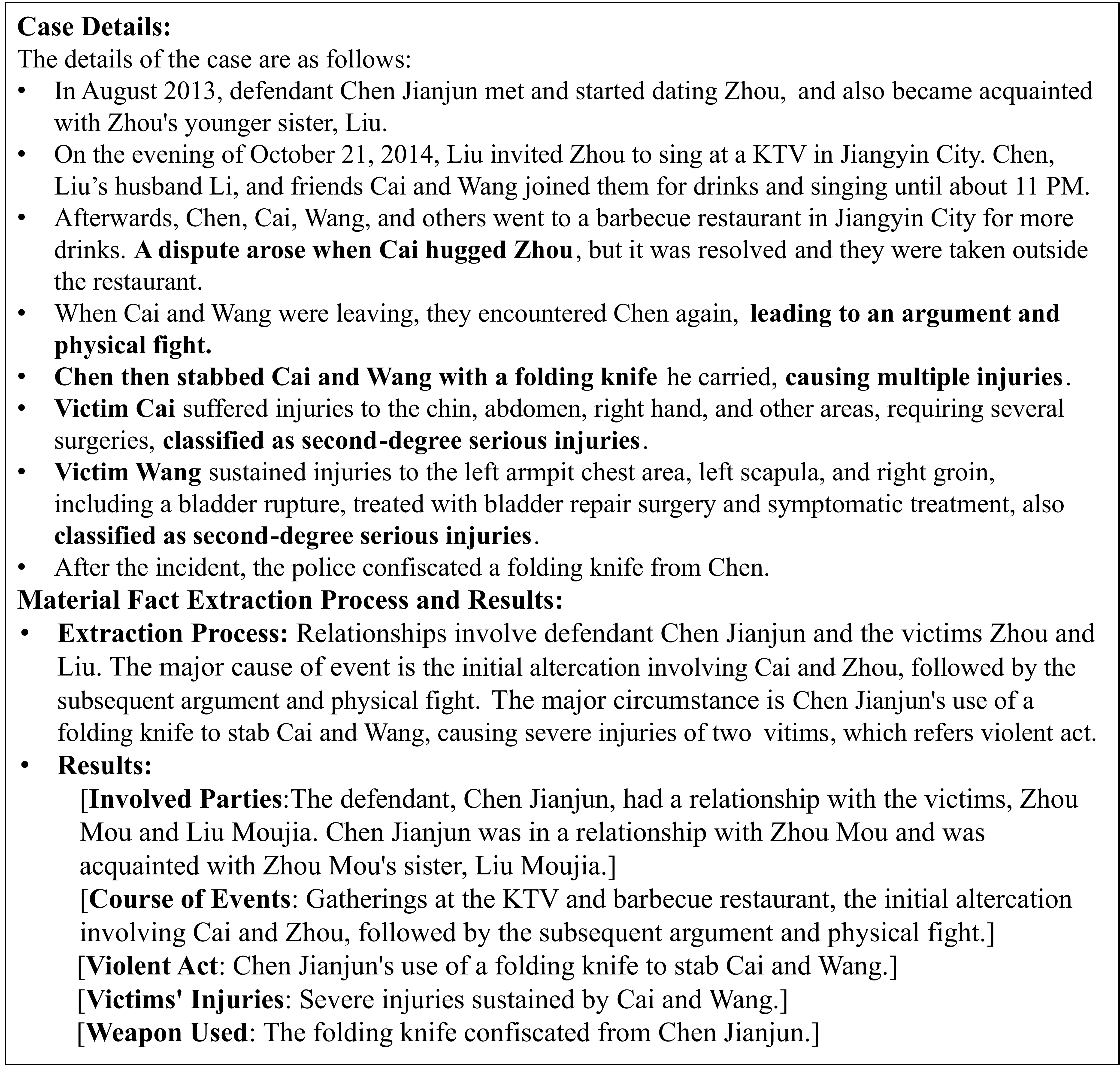} 
	\caption{An example of demonstration used in Fact Extraction (FE) for extracting Material Facts (MF). (Translation of Chinese texts.)}      
	\label{mf}
    
\end{figure}

\begin{figure}[h]  
	\centering
	\includegraphics[width=1\columnwidth]{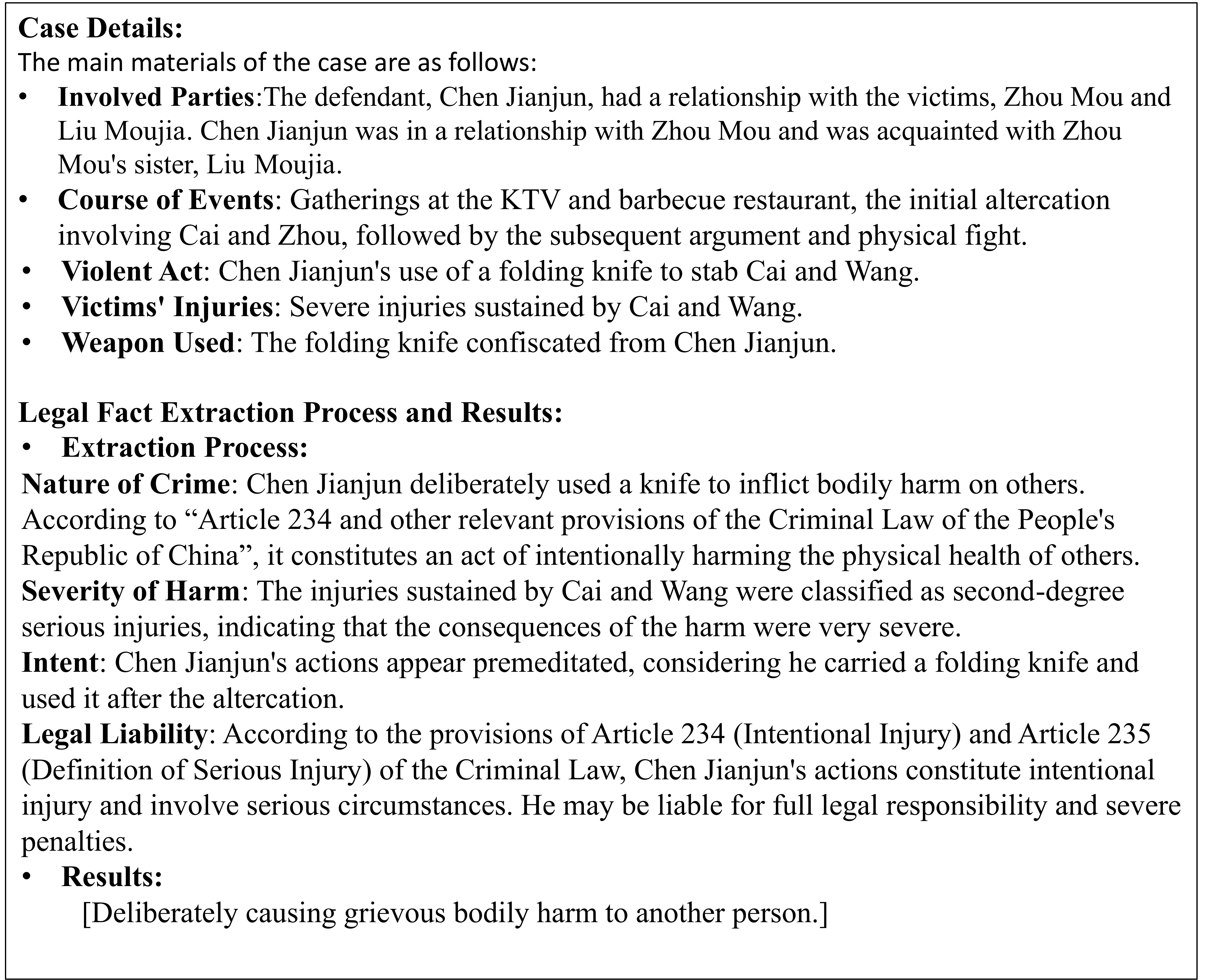} 
	\caption{An example of demonstration used in Fact Extraction (FE) for extracting Legal Facts (LF). (Translation of Chinese texts.)}      
	\label{lf}
\end{figure}

\textbf{Preliminary factual analysis by legal professionals}: While the relevance judgments of the original data were made based on the criteria mentioned in Section 3.1, the annotated data only contain relevance scores without explicit matching labels, preventing the understanding of the exact facts that determine the similarity between two cases. Accurate and direct indications of what the 'Material Fact' in the current case is, its connection to the details in the text, followed by additional legal knowledge and reasoning logic used for identifying such connections, can motivate LLMs to utilize their inherent knowledge and comprehension abilities to mimic human experts' cognitive processes. Such an approach is deemed necessary because we observed that, without detailed relevance indications, the model's judgments tend to deviate and exhibit less explanation consistency. For instance, if case 1 involves person A stealing a car and case 2 involves person B stealing a wallet, the LLM sometimes perceives a car and a wallet as dissimilar items, focusing on trivial facts, thereby simply judging the two cases as irrelevant. 
Thus, we consider such legal reasoning order as a logical chain, instructing the LLM to make judgments consistent with those of the professionals. Figure \ref{mf} and Figure \ref{lf} show examples of experts' demonstrations in FE for MF and LF.

\begin{figure}[h]  
	\centering
	\includegraphics[width=1\columnwidth]{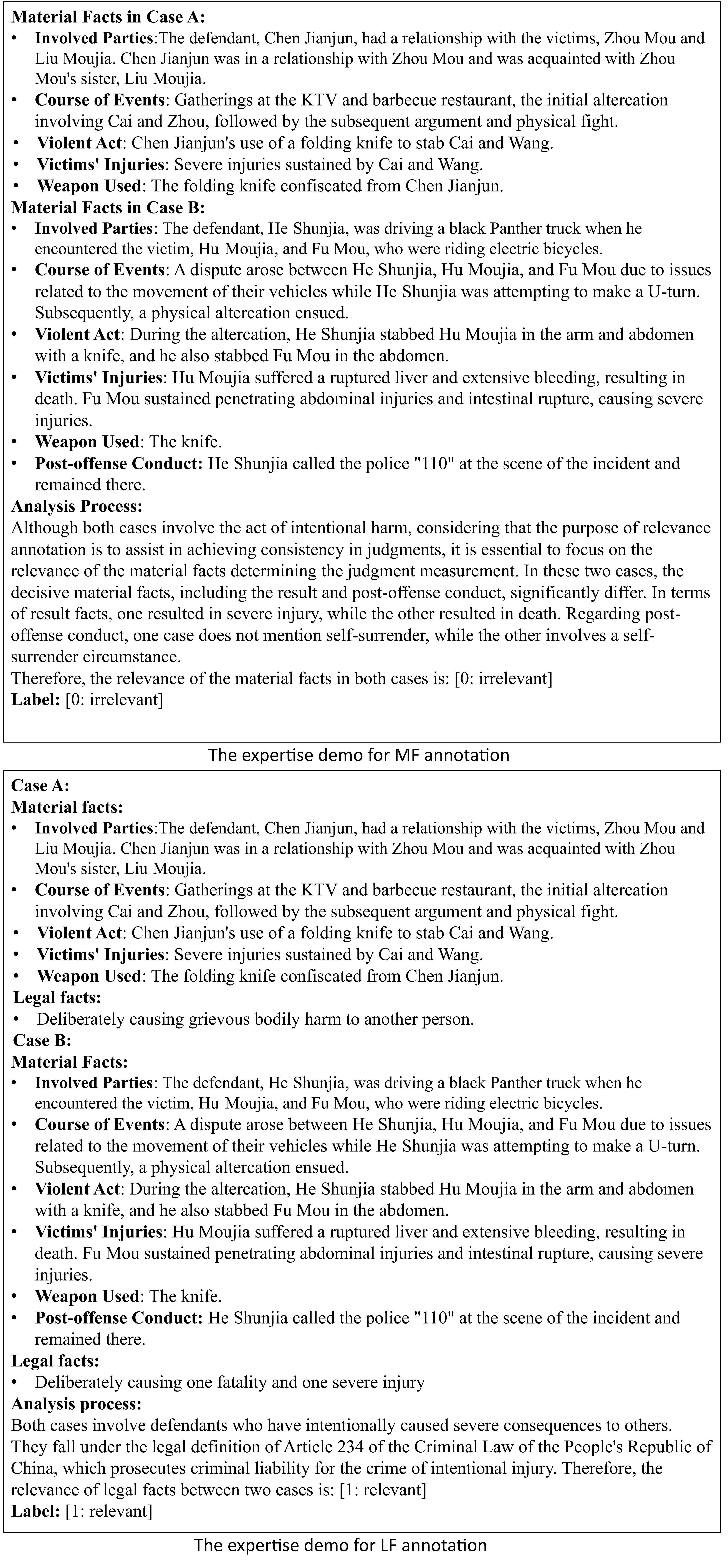} 
	\caption{Two examples of demonstration used in Fact Annotation (FA) for judging Material Facts (MF) and Legal Facts (LF) respectively. (Translation of Chinese texts.)}    
	\label{anno}
\end{figure}
As noted in Section 3.1, legal judgments should follow a sequential structure: Material Facts are established first, followed by Legal Facts. Consequently, the Fact Extraction of LF involves a multi-dimensional analysis based on the Fact Extraction of MF, including considerations such as the defendant’s subjective intent, possible contributory fault of the victim, and the causal relationship between the criminal act and the resulting harm. Then, during the Fact Annotation phase, it is crucial to perform adaptive relevance assessments tailored to different case pairs, based on the connection between MF and LF. We incorporate carefully designed expert reasoning logic for relevance analysis to guide the LLM in following experts' judging criteria.

Formally, for a given stage \( t \in \{\text{FE}, \text{FA}\} \) and fact type \( f \in \{\text{LF}, \text{MF}\} \), the expert-curated demonstration sets is denoted as $\mathcal{D}_{t}^{f}$ for a given stage \( t \in \{\text{FE}, \text{FA}\} \) and fact type \( f \in \{\text{LF}, \text{MF}\} \), where

\[
\begin{aligned}
\mathcal{D}_{\text{FE}}^{\text{LF}} &= \{ \text{Expert demonstrations for Legal Facts in Fact Extraction} \}, \\
\mathcal{D}_{\text{FE}}^{\text{MF}} &= \{ \text{Expert demonstrations for Material Facts in Fact Extraction} \}, \\
\mathcal{D}_{\text{FA}}^{\text{LF}} &= \{ \text{Expert demonstrations for Legal Facts in Fact Annotation} \}, \\
\mathcal{D}_{\text{FA}}^{\text{MF}} &= \{ \text{Expert demonstrations for Material Facts in Fact Annotation} \}.
\end{aligned}
\]

Figure \ref{anno} shows an example of demonstrations for the Fact Annotation.


\textbf{Adaptive Demo-Matching (ADM)}: 
Adaptive selection of demonstrations for few-shot in-context learning ensures better alignment between the input and expert instruction, thereby enhancing the LLM’s ability to emulate expert legal reasoning.

Therefore, we use a retrieval function \( R(\cdot) \) to adaptively retrieve case-specific demonstrations, which will be placed into the input prompt template. Specifically, given an input case \( x \), we retrieve the top-\( k \) most relevant examples from the expert-curated demonstration sets. For a given stage \( t \in \{\text{FE}, \text{FA}\} \) and fact type \( f \in \{\text{LF}, \text{MF}\} \), we select demonstrations as follows:

\[
\text{ADM}(x; t, f) = R\left(x; \mathcal{D}_{t}^{f}\right)
\]
where \( R(x; \mathcal{D}) \) retrieves the top-\( k \) most similar examples to \( x \) from \( \mathcal{D} \), using BM25 in our setting.

\begin{figure}[h]  
	\centering
	\includegraphics[width=1\columnwidth]{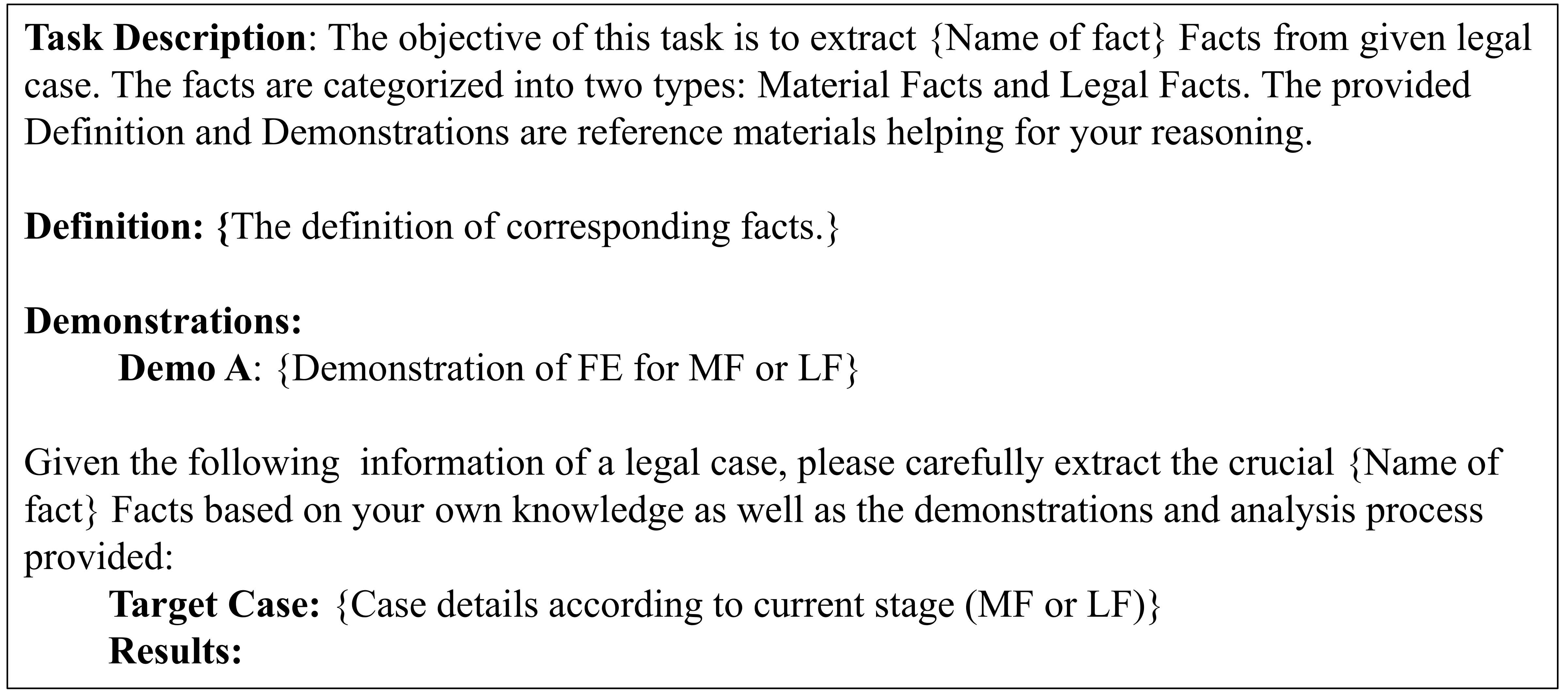} 
	\caption{The example of fact extraction FE. (Translation of Chinese texts.)}      
	\label{fe}
\end{figure}

\textbf{Fact Extraction (FE)}: As illustrated in Figure \ref{fe}, we present an framework of FE prompting. Following the order of 'Material Facts first, Legal Facts second', we separately extract MF and LF sequentially. In the prompt, we provide the task description, definitions of Material Facts or Legal Facts, and the top 2 demos collected by ADM. Finally, the target case is input into the prompt. Specifically, for the MF extraction, the input is the case details, and for the LF extraction is the extracted MF.

\begin{figure}[h]  
	\centering
	\includegraphics[width=1\columnwidth]{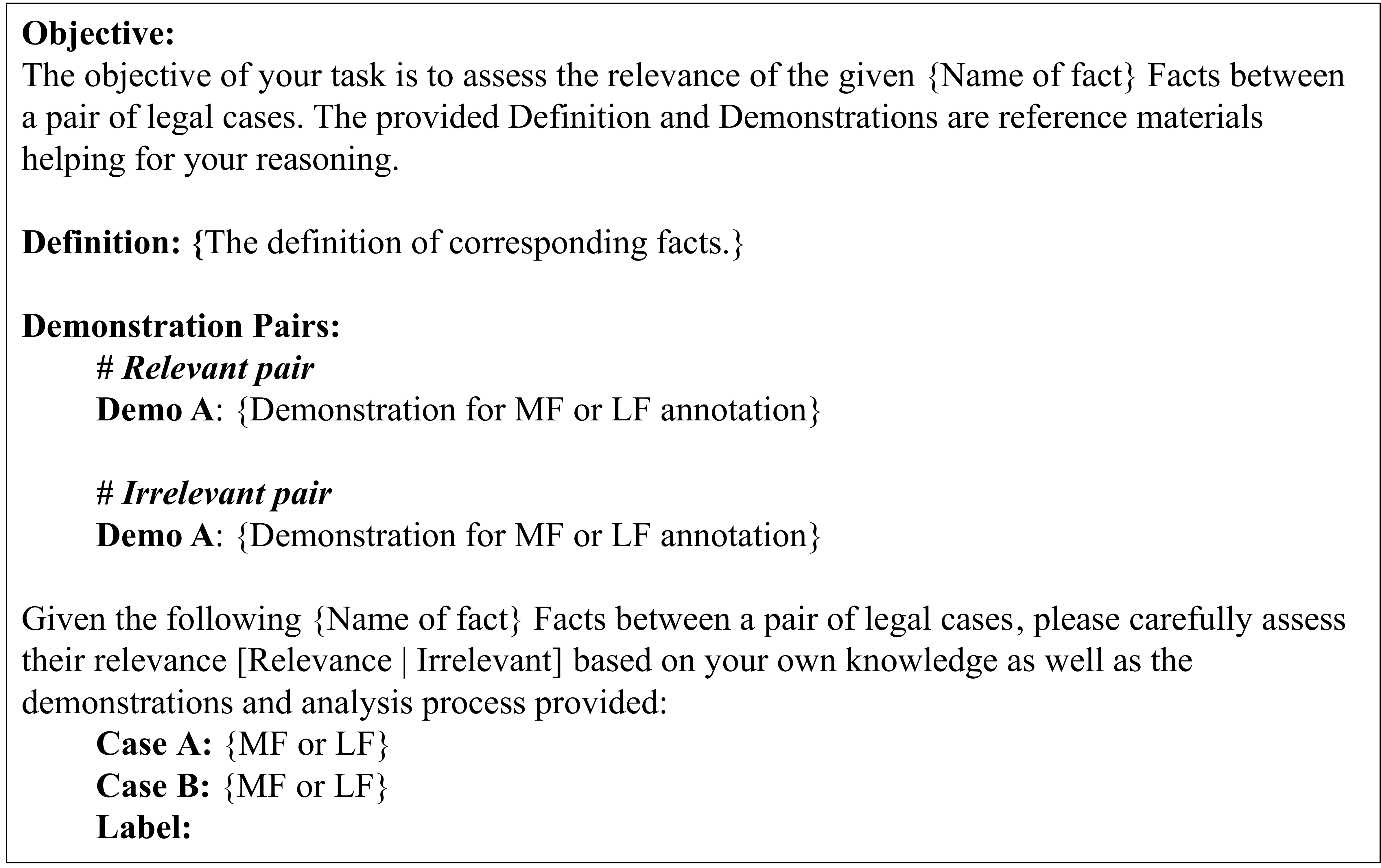} 
	\caption{The example of Fact Annotation (FA). (Translation of Chinese texts.)}      
	\label{fa}
\end{figure}

\textbf{Fact Annotation (FA)}: As illustrated in Figure \ref{fa}, we present an framework of FA prompting. Similarly, we separately annotate MF and LF sequentially. Given as input the facts of a pair of cases to be labeled, the prompt contains relevant as well as irrelevant demonstrations.  Ultimately, the LLMs are instructed to assess the relevance of the current input pair, drawing upon the expert's rationale. The calculation of the final relevance score follows the method of LeCaRD: score 1 for MF relevance, and score 2 for LF relevance, with the total relevance score calculated as the sum of MF and LF relevance scores.

\subsection{Other Statements}
Our work is currently focused on Chinese law due to the easy access to Chinese legal experts, who can help us build a small yet high-quality set of expert demonstrations. Additionally, given the scarcity of annotated legal data (the very issue we aim to address), we conduct research on LeCaRD, the sole case retrieval dataset in the Chinese legal domain. To apply our approach to other countries' legal cases, the same process can be utilized. The only necessary change is to refine the content under the framework of our demonstrations, with the help of domain experts in the corresponding legal field and languages.

Apart from that, the data of Chinese legal documents includes more than just the "basic case information", but we only utilize this part for relevance judgments. There are two main reasons for this:
\begin{itemize}
    \item \textbf{Fact-Driven}: In practical scenarios for judges and lawyers, it is necessary to retrieve similar cases based on the details of new cases that only contain case facts. It's the same in the query cases of the Lecard dataset, which only contain case facts. The goal of legal case retrieval is to ensure similar cases receive fair and consistent adjudication, which is why the analysis starts from the case itself, rather than basing it on the outcomes of judgments. Our approach is thus in line with practical requirements.
    \item \textbf{Consistency with Dataset Standards}: By focusing on case facts, our approach aligns with the Lecard dataset's method of judgments, which is based on key facts (MF and LF) within case details. This ensures that our method aligns with the dataset's standards for relevance annotation.
\end{itemize}

\subsection{Utility of the Relevance Judgments}
To further analyze the utility of our relevance judgments, we use the data annotated with relevance judgments to perform data augmentation for downstream models: to construct a synthetic dataset from unlabeled legal cases for model training. Take LeCaRD as an example, where there are a total of 10,700 candidate cases. By randomly selecting two candidate cases each time for relevance annotation, theoretically, we can generate over 50 million pairs of data. This could significantly scale the annotated dataset. We tried this on a small scale: We first randomly generated 200k candidate case pairs. Our goal is to produce more high-quality positive examples. Considering that the case pairs randomly extracted have a high probability of being irrelevant, the 200k case pairs were pre-sorted by a BERT model fine-tuned on the original dataset (see the Experiment section for implementation details) and we performed relevance annotation on the top 50k data. After obtaining these annotations, we first constructed two synthetic label-only datasets intended for fine-tuning the retrieval models: the first, a 2k and a 20k dataset with distributions that closely mirror the label distribution of the original dataset; and the second, a 40k dataset selected randomly from the 50k annotations. Then we use the 2k synthetic dataset, along with the LLM analysis process saved during its generation, to fine-tune open-source small LLMs, enhancing their capability of legal relevance judgments.

\section{Experiment}
In this section, we conduct various empirical experiments to answer the following research questions: 

\textbf{RQ 1}: Does the workflow reliably produce consistent and reasonable judgments? 

\textbf{RQ 2}: How do the LLM-based relevance judgments align with the judgments from human annotators? 

\textbf{RQ 3}: How can LLM-based relevance judgments be utilized to enhance existing legal case retrieval models and to train open-source large models for assessing case similarity? 

\textbf{RQ 4}: How much does each component of our workflow contribute to the overall judgment accuracy?

Briefly, \textbf{RQ1} focuses on internal workflow consistency and reliability, while \textbf{RQ2} focuses on external alignment with human standards. Both questions are important and complementary, as they address different aspects of the workflow's performance.

\subsection{Datasets and Evaluation Metrics}

We carried out our experiments using the LeCaRD (Chinese Legal Case Retrieval Dataset)~\cite{ma2021lecard}. This dataset comprises more than 43,000 candidate cases and 107 query cases. All the queries and results are sourced from criminal cases made public by the Supreme People's Court of China. Notably, while only the fact paragraph is present in the LeCaRD queries, the candidate documents encompass the full case. For every query, there is a set of 100 corresponding candidate cases. In LeCaRD, relevance is categorized with multi-level labels (0 to 3). For a pair of cases: 0 indicates no relevance, 1 indicates that the Material Facts are relevant but the Legal Facts are not, 2 indicates that the Legal Facts are relevant but the Material Facts are not, and 3 indicates relevance to both kinds of facts. 

To the best of our knowledge, our method is the first for automatic Chinese legal case relevance judgments. To evaluate our method, we compare the judgment consistency with the human-annotated labels by Cohen's Kappa.
In addition, we choose normalized discounted cumulative gain (NDCG) for evaluating the performance of legal case retrievers before and after training on the augmented data, to indirectly illustrate the effectiveness of our method.

\subsection{Implementation Details}
According to the crime classification defined in the chapters of the Criminal Law's specific provisions, the expert-curated demonstration set for ADM is constructed based on 40 query cases, which cover the main types of cases present in the LeCaRd dataset. For each query case, four associated cases with different gold-standard relevance labels are provided, along with FE and FA instructions corresponding to two factual aspects: MF and LF.

For the relevance evaluation, we compare the judgments of the proposed method with the golden labels of the top 30 candidate cases in every query case. For the LLM, we opted for GPT-3.5-turbo with temperature 0.4. In the heatmap experiment, we randomly sample 400 human-labeled pairs for all relevance levels, from 0 to 3. While more capable models (e.g., gpt-4o) were available in later stages of this research, we retained gpt-3.5-turbo for all experiments to maintain comparability with earlier results. Given that gpt-4o consistently surpasses gpt-3.5-turbo in reasoning and generation quality, we expect our method’s performance to further improve when paired with stronger backbones, which will be investigated in future studies.

For the data augmentation experiments, we split LeCaRD in a ratio of 8:1:1 for training, validation, and testing. The augmented data pairs are sampled from the candidate case corpus from the training set. We use the Chinese tokenizer Jieba~\footnote{https://github.com/fxsjy/jieba} in data pre-processing and use AdamW~\cite{loshchilov2017decoupled} optimizer in fine-tuning retrieval models. All hyper-parameters are tuned based on the performance of the validation set. Specifically, when using the BERT~\footnote{https://huggingface.co/google-bert/bert-base-chinese} baseline model, we adopted a segmentation method similar to BERT-PLI, splitting long texts into chunks of length 256 with overlapping window sizes of 128. The [CLS] token serves as the embedding for each block, and during relevance computation, we cross-calculate the dot product of the query and candidate block vectors. Furthermore, the LeCaRD is in Chinese, so we use the Chinese version of Longformer, longformer-chinese. For the LLMs, we set the temperature to 0.5. When fine-tuning open-source LLMs using 2k synthetic data of GPT-3.5 with its analysis process, we applied LoRA with a batch size of 256, a rank of 32, a learning rate of 5e-4, and a context window of 2048.

\subsection{Evaluation of the Relevance Judgments (\textbf{RQ1 \& 2}) }
\begin{figure}[h]
    \centering
    \includegraphics[width=0.75\columnwidth]{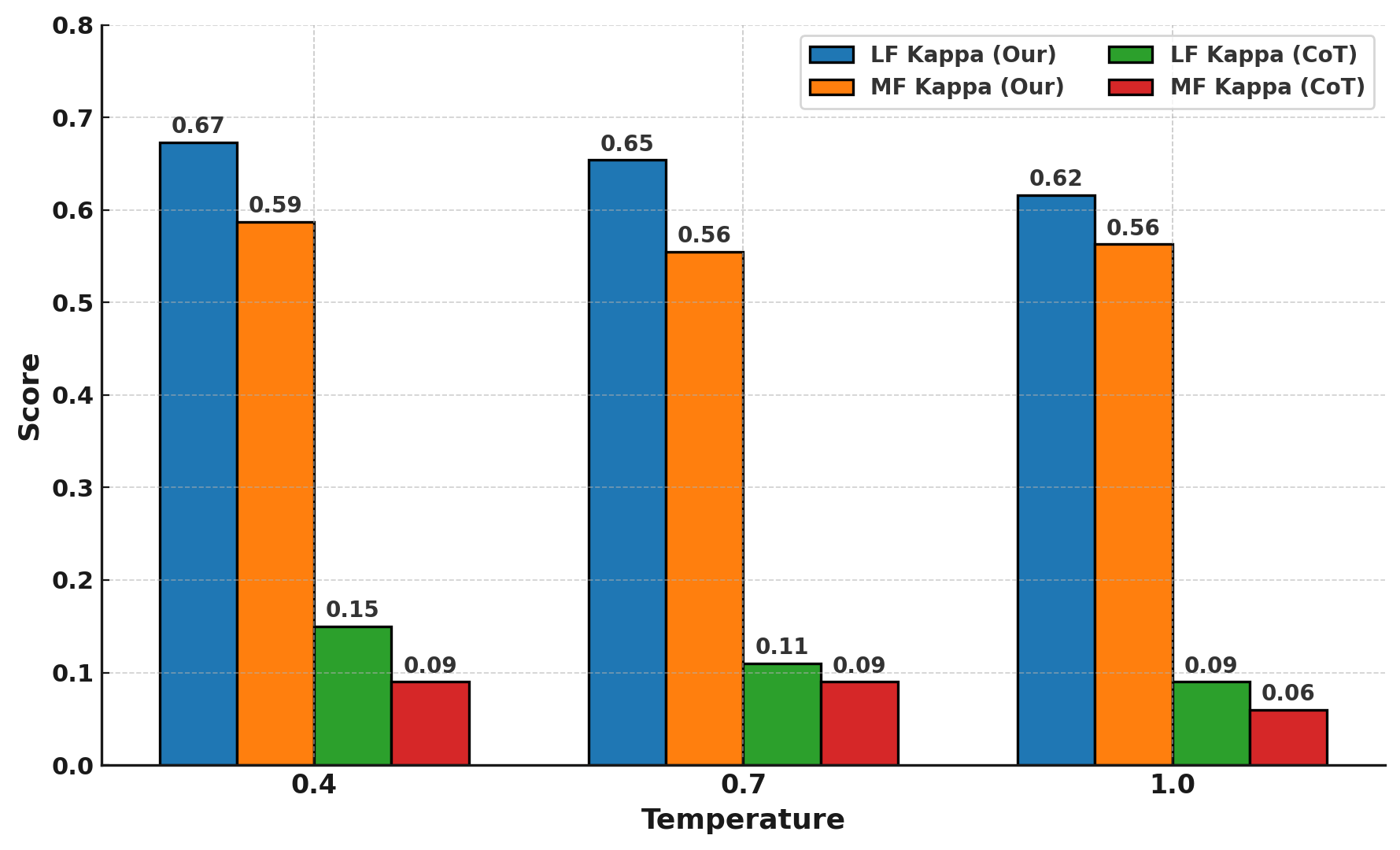}
    \caption{Reliability Test: The Cohen's Kappa between multiple times of MF\&LF relevance judgment over different temperature settings.}
    \label{kappa2}

\end{figure}

\begin{figure}[h]
    \centering
    \includegraphics[width=0.75\columnwidth]{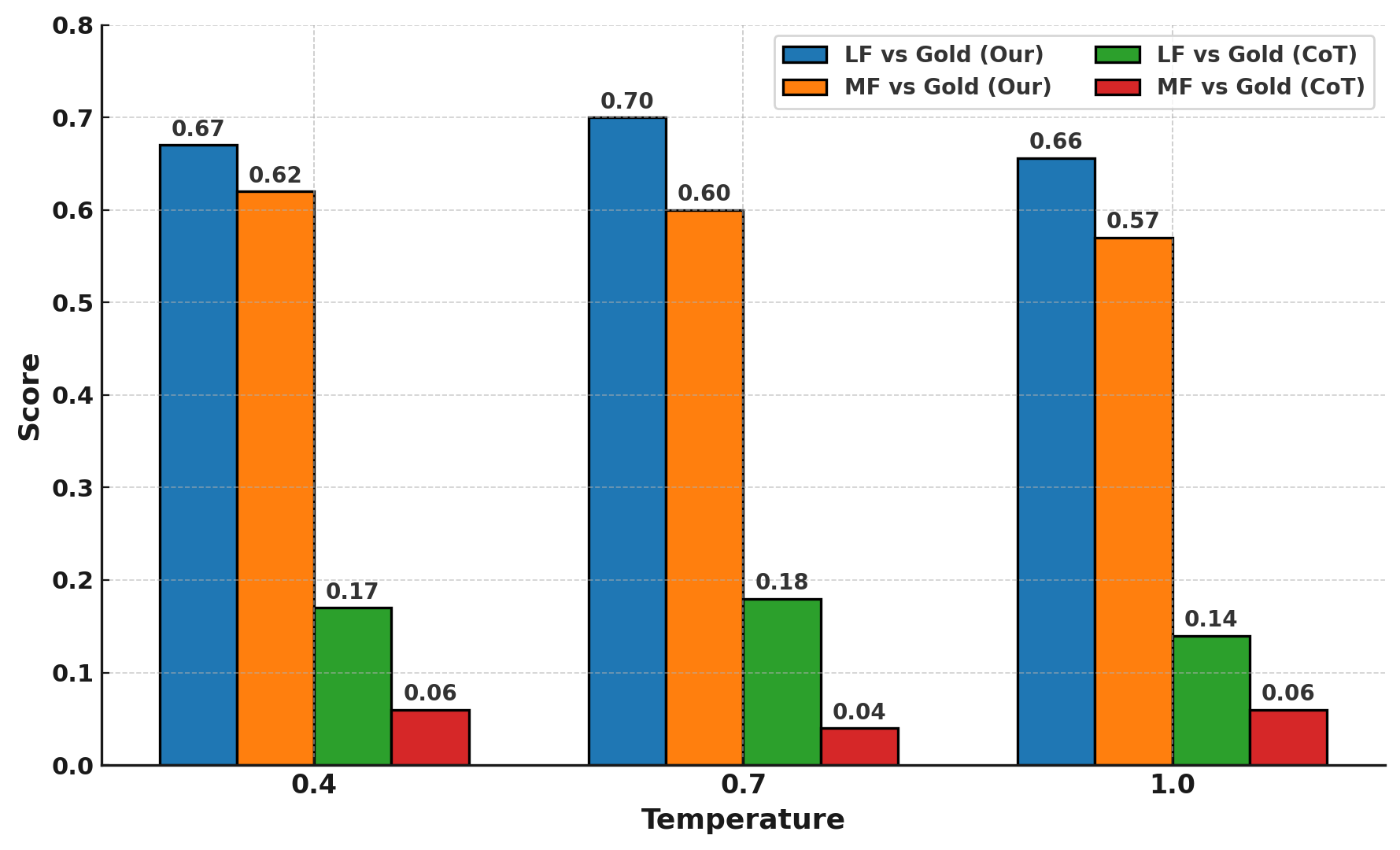}
    \caption{Validity Test: The Cohen's Kappa between the MF\&LF and the golden labels over different temperature settings: Avg. represents the average of Kappa values between multiple relevance judgments and the golden labels}
    \label{kappa1}
\end{figure}

Separately based on two aspects, Material Facts (MF) and Legal Facts (LF), we assess the reliability of the proposed method by calculating Cohen's Kappa for multiple judgments conducted at a certain temperature in Figure~\ref{kappa2}. To demonstrate the validity, we evaluate the automated judgment consistency by calculating Kappa values with the golden labels in Figure~\ref{kappa1}, and then draw heatmaps that visualize the consistency between the proposed method and human labeling. To more intuitively demonstrate the reliability and validity of our method, we also report relevance judgment results obtained directly using a Chain-of-Thought (CoT) prompting strategy（with all necessary labeling guidance and few shot demonstrations）.

For the validity test in Figure~\ref{kappa1}, the consistency between Legal Facts and human labels is highly significant. This demonstrates that our automatic relevance judgment closely approximates human experts' labeling. We also observe that Material Facts show only relative significance. A key reason is that each case involves one or more defendants; when comparing defendant-related facts between cases, one-to-many and many-to-many relationships often require subjective judgment. We observed such subjectivity in human annotations of Material Facts, partly because labels come from multiple annotators with inevitable individual biases. Moreover, this bias is more likely to manifest in Material Fact, due to the arbitrary forms of writing it. Nonetheless, according to the labeling standard of LeCaRD, Legal Facts (score=2) are more critical than Material Facts (score=1), so the overall quality of Legal Fact judgments dominates final results. \textbf{Notably, directly employing a CoT prompting strategy for case relevance judgment yields poor performance in both reliability and validity, underscoring the necessity of our approach.}

\begin{figure}[h]  
	\centering
	\includegraphics[width=0.7\columnwidth]{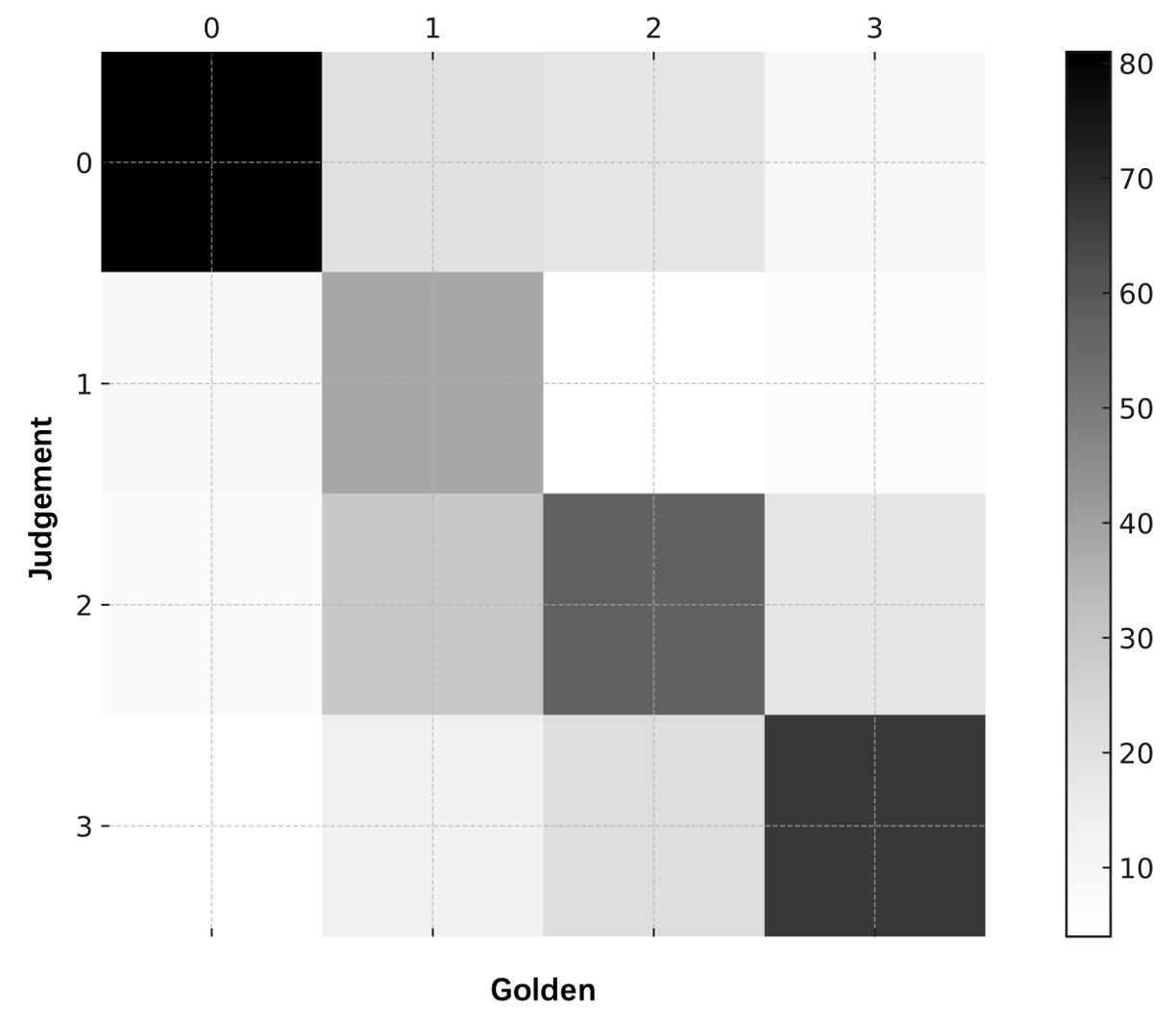} 
	\caption{The heatmap of the four-level relevance judgments vs golden labels.}      
	\label{hm}
\end{figure}

\begin{figure}[h]  
	\centering
	\includegraphics[width=0.95\columnwidth]{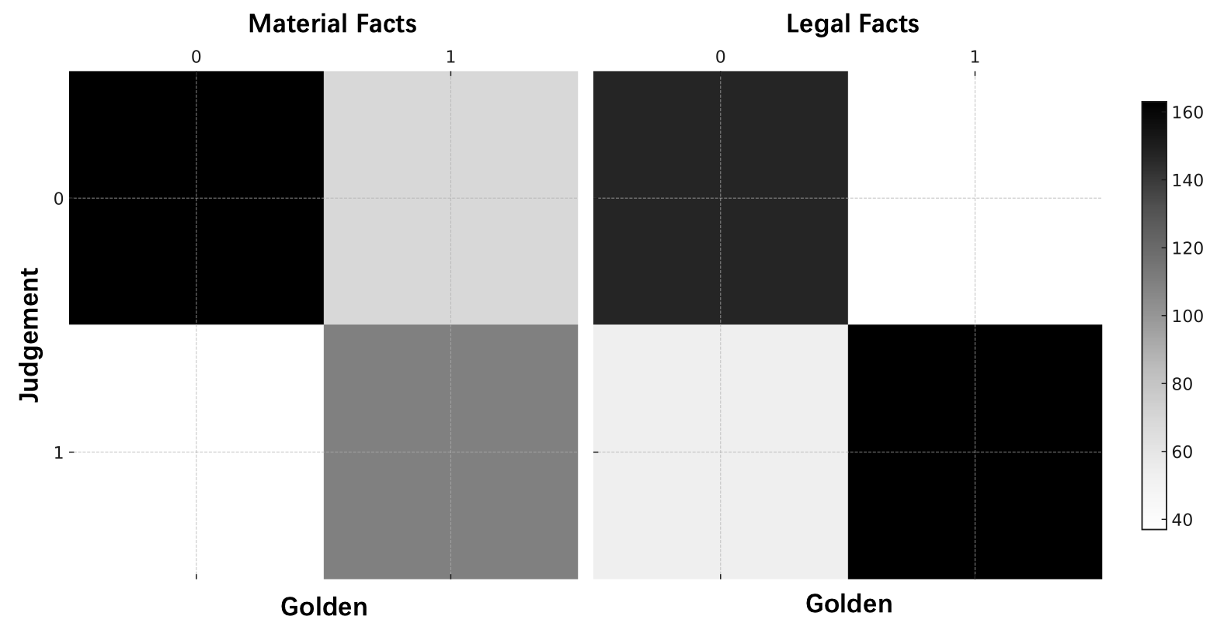} 
	\caption{The heatmap of the relevance judgments MF\&LF relevance vs golden labels.}      
	\label{hm2}
\end{figure}

The heatmaps in Figure~\ref{hm} and Figure~\ref{hm2} show the high consistency of our automated relevance judgments with human labeling, especially in cases of non-relevance and full relevance. This further demonstrates the validity of the judgment results.

\subsection{Data Augmentation Experiments (\textbf{RQ3})}
To explore the annotation capabilities of different LLMs and whether the proposed workflow in this paper can further enhance the annotation capabilities of open-source light-weighted LLMs through fine-tuning. We first fine-tune various small LLMs using generated 2k synthetic data with analysis processes annotated by GPT-3.5. Then, using the proposed workflow with GPT-3.5 replaced by smaller LLMs, we compare their consistency with human-annotated golden labels on the test set.
To further validate the utility of our automatic relevance judgment, we analyze the improvements achieved by training different retrieval models on annotated synthetic data in LCR tasks. 

\subsubsection{Baselines}
For relevance judgment capabilities of different LLMs, we opt for GPT-3.5-0614 (referred to as \textbf{GPT-3.5}), Llama-2-7B-chat-hf\cite{touvron2023llama2openfoundation} (referred to as \textbf{Llama-2}) and Qwen-2-7B-instruct\cite{bai2023qwentechnicalreport} (referred to as\textbf{ Qwen-2}). 

In data augmentation experiments, we select two basic deep models (\textbf{BERT}~\cite{devlin2018bert}, \textbf{Longformer}~\cite{beltagy2020longformer}) and an advanced SOTA method \textbf{SAILER}~\cite{li2023sailerstructureawarepretrainedlanguage} as our baselines. 
\textit{\textbf{Note that our primary objective is to assess the quality of the automatic relevance judgments. We evaluate this by measuring the relative performance gains in downstream models after applying the augmented data, rather than aiming to achieve a SOTA model on the LCR task.}}


\subsubsection{Data Augmentation Results for Fine-tuning small LLMs} 
As shown in Table~\ref{opensource_llms}, using our proposed workflow and a small amount of synthetic data generated by the powerful general LLM GPT-3.5, the capability for legal relevance judgment of two 7B LLMs can be significantly improved, reaching or even surpassing that of GPT-3.5. 

\begin{table}[ht]
\centering
\small
\setlength{\tabcolsep}{4pt} 
\renewcommand{\arraystretch}{1} 
\caption{Relevance consistency (Cohen’s Kappa) with golden labels in the test set, where * denotes models fine-tuned with synthetic data.}
\begin{tabular}{lccccc}
\toprule
  & \textbf{GPT-3.5} & \textbf{Llama-2} & \textbf{Qwen-2} & \textbf{Llama-2\textsuperscript{*}} & \textbf{Qwen-2\textsuperscript{*}} \\
\midrule
LF & 0.67 & 0.41 & 0.44 & 0.63 & \textbf{0.69} \\
MF & 0.62 & 0.25 & 0.29 & 0.55 & 0.60 \\
\bottomrule
\end{tabular}
\label{opensource_llms}
\end{table}

\subsubsection{Data Augmentation Results for Fine-tuning Retrieval Models}
\begin{table}[h]
\centering
\caption{NDCG@30 of different baselines after pre-training on label-only synthetic data of various sizes, generated by either the original or fine-tuned LLMs. G, L, and Q denote GPT-3.5, Llama-2, and Qwen-2, respectively, while * indicates fine-tuned LLMs with explanations from GPT-3.5. Bold numbers indicate the best performance in each row, while underlined numbers represent the second best. The 2k and 20k datasets have a similar positive-negative ratio to the training set, while the 40k dataset was randomly collected from all generated data.}
\resizebox{\columnwidth}{!}{ 
\begin{tabular}{lcccccc}
\toprule
\textbf{Baseline}     & \textbf{Original} & \textbf{G} & \textbf{L} & \textbf{Q} & \textbf{L\textsuperscript{*}} & \textbf{Q\textsuperscript{*}} \\
\midrule
\textbf{BERT}                  & 0.8816 & -       & -       & -       & -       & -       \\
\textbf{Longformer}            & 0.9019 & -       & -       & -       & -       & -       \\
\textbf{SAILER}            & 0.9223 & -       & -       & -       & -       & -       \\
\midrule
\textbf{BERT}+2k               & -      & \textbf{0.8979} & 0.8794 & 0.8851 & 0.8895 & \underline{0.8967} \\
\textbf{BERT}+20k              & -      & \underline{0.9305} & 0.9043 & 0.9156 & 0.9241 & \textbf{0.9330} \\
\textbf{BERT}+40k              & -      & \textbf{0.9156} & 0.8962 & 0.9007 & 0.9087 & \underline{0.9125} \\
\textbf{Longformer}+20k        & -      & \underline{0.9458} & 0.9129 & 0.9283 & 0.9374 & \textbf{0.9480} \\
\textbf{SAILER}+20k        & -      & \underline{0.9606} & - & - & 0.9479 & \textbf{0.9623} \\
\bottomrule
\end{tabular}
}
\label{main}
\end{table}

\begin{table}[h]
\centering
\caption{The impact of different components for relevance quality: Original indicates the proposed approach. w/o ADM indicates the method of selecting examples from the demo library using random sampling; w/o FE means letting the LLM extract the two types of facts directly in the prompting workflow without prompt guidance and demo examples; w/o FA indicates that after extracting the two types of facts from each case, the similarity is directly labeled without the guidance of positive and negative demos in the original FA progress. }
\begin{tabular}{lcl}
\toprule
Methods & MF Kappa & LF Kappa\\
\midrule
Original &  0.62 & 0.67\\
w/o ADM &  0.45 & 0.58\\
w/o FE & 0.21 & 0.30\\
w/o FA &  0.26 & 0.38\\
w/o FE\&FA &  0.19 & 0.26\\
\bottomrule
\label{aba}
\vspace{-0.7cm}
\end{tabular}
\end{table}


First, in a vertical comparison presented in Table~\ref{main}, BERT trained only on the 2k synthetic dataset already shows significant improvements. Performance further improves substantially with the larger 20k dataset, indicating that our method can make effective relevance judgments. The improvement on the 20k synthetic data is shown across all baselines, demonstrating the general effectiveness of the synthetic supervision of our method. Notably, both BERT and Longformer, when trained on the 20k synthetic dataset, outperform the original SOTA model SAILER, highlighting the strong potential of our approach despite not relying on any additional complex post-processing techniques.
Interestingly, despite doubling the data size, the performance on the 40k dataset was noticeably worse than on the 20k dataset. Two potential explanations are as follows: (1) a training dataset more aligned with the test set distribution yields greater benefits for the model; and (2) random sampling without filtering or selection introduces more unnecessary negatives and fewer positive examples, with excessive negatives adding limited value to the model.   
Then, in a horizontal comparison, we observed a phenomenon similar to that in Table 3: small LLMs fine-tuned on data augmented by a strong LLM using our workflow, when further used to annotate data within the same workflow for fine-tuning retrievers, achieve improvements that can approach or even surpass the performance of the strong LLM, With minimal expert supervision, our approach enables a large language model to acquire case analysis expertise, and subsequently transfers this ability to a smaller model via annotation-based knowledge distillation.

\subsection{Ablation Study (\textbf{RQ4})}
In this section, we aim to explore the impact of each stage in the proposed workflow on the final accuracy of the relevance judgments.  


According to Table~\ref{aba}, removing Adaptive Demo-Matching (ADM) impacts the relevance judgments of Material Facts (MF) notably. This indicates that the extraction and relevance assessment of MF relies on more pertinent expert guidance. This may be due to the variable nature of MF. Meanwhile, Legal Facts (LF) remain unaffected due to their normative formats. This also highlights that the understanding of LF in the case benefits more from the Fact Extraction(FE) and Fact Annotation (FA) modules. When FE is removed, the Kappa scores for both types of facts significantly decrease, and the effect is similar when both FE and FA are removed. This demonstrates that the fact extraction step we designed is the most crucial part of the entire judgment process. Based on this observation, it can be inferred that LLM faces difficulty in accurately identifying key legal elements. Removing FA has little impact on LF annotation, but it does negatively affect MF. This indirectly confirms that extracting MF from case pairs and assessing their relevance is a challenging aspect of judging legal case relevance, and our method addresses this challenge through a series of carefully designed processes.


Every stage within the workflow contributes significantly to enhancing the quality of legal relevance judgments. Both ADM and FA moderately influence the quality of relevance judgments, with the impact of ADM being more pronounced. ADM validates that relevant demonstrations indeed provide LLMs with a more robust basis for reasoning. On the other hand, FA introduces clearer constraint conditions for relevance judgments. Notably, the most pivotal process is FE, involving the most specialized knowledge Furthermore. It also suggests that our FE method can effectively leverage minimal expert instructions to inspire and guide LLMs in capturing subtle but vital nuances within professional data.

\section{Limitation and additional experiments}
Our study has two main limitations. First, we only evaluate on the case retrieval task, though it is highly relevant in real-world legal scenarios. Second, all experiments are conducted in Chinese, due to substantial differences between Chinese and foreign legal systems.

Despite this, our method generalizes well to other legal domains. We conducted additional experiments on CAIL2019\footnote{\url{https://github.com/alumik/cail2019}}, which involves civil cases such as private lending, IP disputes, and maritime law. Unlike LeCaRD, each sample contains a query case and two candidates, requiring a binary relevance judgment.
\begin{table}[h]
\centering
\caption{Results for CAIL2019 dataset}
\vspace{-5pt}
\begin{tabular}{@{}cccc@{}}
\toprule
Model       & SOTA  & BERT  & BERT+20k \\ \midrule
F-score     & 0.739 & 0.436 & \textbf{0.750} \\
\bottomrule
\end{tabular}
\vspace{-5pt}
\label{666}
\end{table}
Using only 9 expert-designed demonstration sets, we generated pairwise annotations from unlabeled cases in the training set and selected 20k samples to fine-tune BERT. As shown in Table\ref{666}, results compared with the reported SOTA result\cite{feng-etal-2024-legal, bi2022learning} show that our method remains effective in different legal domains.

\section{Conclusions and Future Works}
In this paper, we proposed an automated relevance judgment method for legal cases. Through empirical experiments, it demonstrates good reliability and high consistency with human labels. Using the generated synthetic data can further make a significant improvement for downstream models.

 



\bibliographystyle{ACM-Reference-Format}
\bibliography{sample-base}

\end{CJK}
\end{document}